\documentclass{article}

\usepackage{PRIMEarxiv}

\usepackage[utf8]{inputenc} 
\usepackage[T1]{fontenc}    
\usepackage{hyperref}       
\usepackage{url}            
\usepackage{booktabs}       
\usepackage{amsfonts}       
\usepackage{nicefrac}       
\usepackage{microtype}      
\usepackage{lipsum}
\usepackage{fancyhdr}       
\usepackage{graphicx}       
\usepackage{amsmath}
\graphicspath{{media/}}     

\title{Creation of Novel Soft Robot Designs using Generative AI}

\author{
  Wee Kiat Chan\textsuperscript{1}, Pengwei Wang\textsuperscript{1}, Raye Chen-Hua Yeow\textsuperscript{2} \\
\textsuperscript{1}Department of Mechanical Engineering, National University of Singapore\\
\textsuperscript{2}Department of Biomedical Engineering, National University of Singapore\\
}

\begin{document}
\maketitle

\begin{abstract}
Soft robotics has emerged as a promising field with the potential to revolutionize industries such as healthcare and manufacturing. However, designing effective soft robots presents challenges, particularly in managing the complex interplay of material properties, structural design, and control strategies. Traditional design methods are often time-consuming and may not yield optimal designs. In this paper, we explore the use of generative AI to create 3D models of soft actuators. We create a dataset of over 70 text-shape pairings of soft pneumatic robot actuator designs, and adapt a latent diffusion model (SDFusion) to learn the data distribution and generate novel designs from it. By employing transfer learning and data augmentation techniques, we significantly improve the performance of the diffusion model.    
These findings highlight the potential of generative AI in designing complex soft robotic systems, paving the way for future advancements in the field.
\end{abstract}

\keywords{Soft Robot \and Diffusion \and Generative AI}

\section{Introduction}
Soft robotics has emerged as a compelling field with the potential to transform
industries such as healthcare, manufacturing, and exploration. Inspired by natural
organisms, soft robots offer unique advantages including increased flexibility, adaptability to complex environments, and enhanced safety during human interactions.
Nonetheless, designing effective soft robots presents challenges, particularly in managing the complex interplay of material properties, structural design, and control
strategies.
A key obstacle in soft robot design is the laborious and often iterative nature of
the design process. \cite{pinskier_bioinspiration_2022} Traditional methods rely heavily on manual or semi-automated
approaches, leading to time-consuming processes that may not always yield optimal
designs. Meanwhile, the development of diffusion models in generative AI shows
promising potential due to their high fidelity and flexibility in generating realistic
samples. While initially applied to text-to-image generation, these techniques have
rapidly expanded into the domain of 3D model generation.
In this project, we explore the use of generative AI to create 3D models of soft
actuators, showcasing promising results in terms of fidelity and conceptual under-
standing. However, practical adaptation of these models still requires significant
advancements.

\section{Related Works}
\label{sec:related}

This section outlines various approaches that have been explored to design and optimize soft robotic systems. The section also explores various diffusion models for three-dimensional shape generation and the use of 3D diffusion model for soft robot design. 

\subsection{Soft Robot Design}
Soft robot actuator design approaches can be broadly categorized into four areas: intuitive manual design, model-based design optimization (MBDO), topology optimization (TO), and generative evolutionary design \cite{pinskier_bioinspiration_2022}. Intuitive manual design leverages a designer's creativity for rapid prototyping of novel concepts \cite{ilievski_soft_2011}. Bio-inspired variations of this method focus on emulating biological principles, achieving impressive results in specific applications but potentially lacking generalizability \cite{lee_soft_2017}\cite{mazzolai_soft-robotic_2012}.

MBDO integrates computational models with optimization algorithms to refine designs iteratively. Due to the limited number of design parameters considered and their specificity to a single design iteration, these models are primarily excels for predicting the impact of minor design modifications through simulation. While exhibiting good concordance with experimental results, their generalizability is constrained, restricting their application to new design predictions \cite{pinskier_bioinspiration_2022}\cite{tawk_finite_2020}. Topology optimization (TO) focuses on optimizing material distribution within a design space. While effective for creating lightweight and efficient designs, traditional TO approaches struggle with soft robots due to their complex behavior and intricate internal structures. Furthermore, TO may generate designs that are difficult to manufacture or implement in real-world scenarios \cite{sigmund_topology_2013}. 

Generative and evolutionary design methods address these shortcomings by utilizing heuristic algorithms inspired by nature to explore and optimize designs\cite{eiben_evolutionary_2015}. These methods, including genetic algorithms and graph-based encodings, can coevolve the morphology and control of robots, enabling the development of complex behaviors. While these methods excel at autonomously exploring design spaces and uncovering innovative solutions, challenges such as the simulation-to-reality gap and high computational complexity hinder their direct application in physical robots \cite{pinskier_bioinspiration_2022}.

Overall, generative and evolutionary design methods hold significant promise for the future of soft robot actuator design due to their ability to explore a vast design space and unearth creative solutions that may not be apparent through traditional manual design processes. However, addressing the limitations of simulation accuracy and computational intensity remains crucial for their wider adoption in real-world soft robotic applications.

\subsection{Diffusion Models}
    
Diffusion models in generative AI offer a promising approach for designing soft robots due to their ability to generate high-quality, diverse samples and capture complex, multimodal distributions. These models, such as the Diffusion Probabilistic Models (DPM), have demonstrated state-of-the-art performance in image generation tasks by iteratively refining a noise vector to generate realistic samples. In the context of soft robotics, diffusion models could be leveraged to generate diverse designs with varying material properties, shapes, and functionalities, facilitating the exploration of a vast design space.

One key advantage of diffusion models is their ability to model complex, high-dimensional data distributions efficiently. Unlike traditional Generative Adversarial Networks (GANs) or Variational Autoencoders (VAEs), diffusion models do not require explicit latent variable sampling during training, leading to more stable training and better sample quality. Additionally, diffusion models can capture long-range dependencies in the data, which is crucial for generating complex soft robot designs that exhibit intricate structures and functionalities.

Furthermore, diffusion models offer a principled framework for uncertainty estimation, which is essential for soft robotics applications where designs must be robust to variations in material properties and environmental conditions. By quantifying uncertainty, diffusion models can help designers make informed decisions about the trade-offs between design complexity, performance, and manufacturability. Overall, diffusion models present a compelling avenue for soft robot design, offering advanced capabilities for generating diverse, high-quality designs and quantifying uncertainty in the design process.

Several 3D diffusion models have been proposed, each offering unique features and capabilities for soft robot design. Text2Shape \cite{chen2018text2shape} leverages text descriptions to generate 3D shapes, making it useful for creating 3D shapes based on textual specifications. SDFusion \cite{cheng2022sdfusion} focuses on generating shapes represented by Signed Distance Functions, allowing for the creation of complex, multi-part designs. DreamFusion \cite{poole2022dreamfusion}  uses a pretrained 2D text-to-image diffusion model to perform text-to-3D synthesis to generate 3D shapes. PointE \cite{nichol2022pointe}, likewise, uses a text-to-image diffusion model to create single synthetic view, which then generates a 3D point cloud using a second diffusion model based on the generated image. These 3D diffusion models offer diverse approaches to soft robot design.

\subsection{Diffusion for Soft Robot Design}

DiffuseBot is one of the few research available on the creation of soft robot design using diffusion models \cite{wang2023diffusebot}. DiffuseBot addresses soft robot design challenges by proposing a novel framework that leverages diffusion models and physics simulation for co-designing morphology and control. It achieves this by: (1) 3D shape generation with diffusion-based models - utilizes pre-trained 3D diffusion models to generate diverse and complex robot geometries, (2) robotizing 3D shapes from diffusion samples - converts the generated 3D point cloud data into a solid mesh suitable for physics simulation, (3) optimizing embeddings for improved physical utility - incorporates techniques to optimize the embeddings that condition the diffusion model, prompting the sampling distribution towards generating higher-performing robots based on the physics simulation and (4) co-designing morphology and control - reformulates the diffusion process to incorporate co-optimization over structure and control by leveraging differentiable physics simulation.

DiffuseBot showcases its effectiveness through simulations of robots accomplishing diverse tasks like balancing, landing, crawling, and object manipulation. Additionally, it presents a real-world 3D printed robot as a proof-of-concept. While impressive, the paper acknowledges limitations in its material property and actuator placement definition. Specifically, DiffuseBot adopts a simplified approach for actuation by embedding "muscle fibers" within the robot's solid geometry. These fibers contract or expand to induce deformation, and a constant stiffness parameterization governs the relationship between this deformation and the resulting restorative elastic force. Overall, DiffuseBot contributes a powerful new method for soft robot design by combining generative power with physics-aware optimization \cite{wang2023diffusebot}.

\section{Method}
\label{sec:method}
We based our approach on SDFusion \cite{cheng2022sdfusion}, adapting it for text-to-shape generation in soft robotics. SDFusion relies heavily on Latent Diffusion, a concept we will discuss in detail later. Our text-to-shape model comprises three main components: an encoder-decoder for transforming objects between the object space (represented as Signed Distance Fields, SDFs) and the latent space, a denoising network for backward diffusion, and a text encoder for conditioning. For the encoder-decoder, we use the Vector Quantized Variational AutoEncoder (VQ-VAE) \cite{van2017neural}, and for the denoising network, we employ a 3D adaptation of the U-Net architecture \cite{ronneberger2015u}. As for the text encoder, we utilize BERT \cite{devlin2018bert}.

\subsection{VQ-VAE}

\begin{figure} [htbp]
    \centering
    \includegraphics[width=1\linewidth]{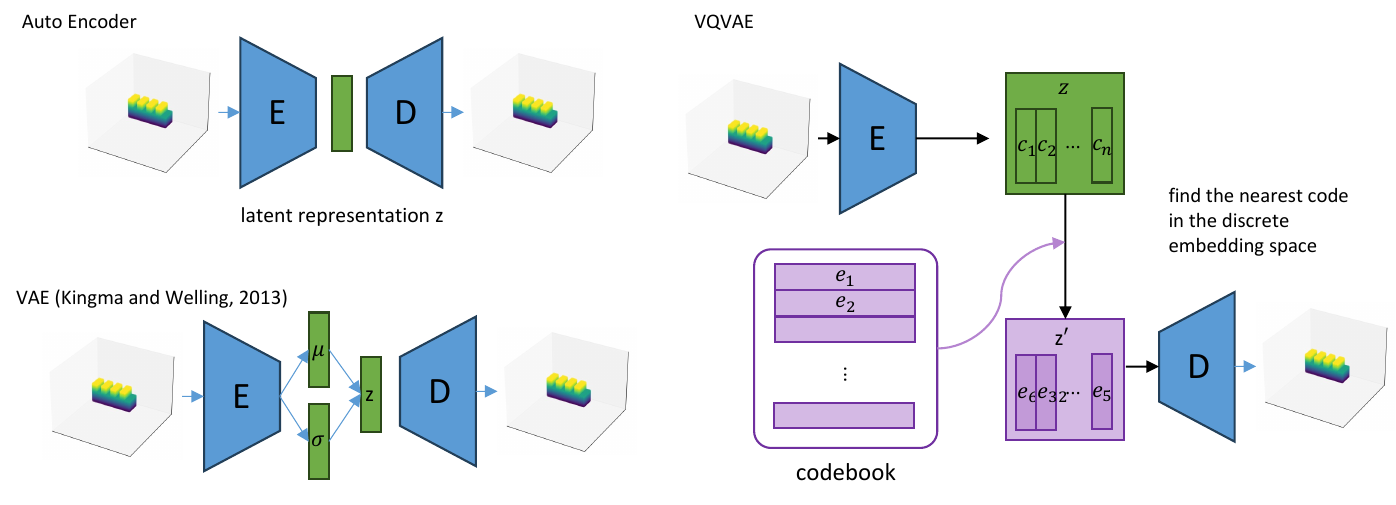}
    \caption{The network comparison between traditional autoencoder, variational autoencoder \cite{kingma2013auto} and vector quantised variational autoencoder \cite{van2017neural}. The picture is drawn by us by referred from the original papers.}
    \label{fig:vqvae-net}
\end{figure}

AutoEncoders (AE) are a kind of neural networks to learn a compressed representation for a given data distribution. A encoder \(f_e\) is used to compress data \(x\) into latent representation \(z = f_e(x)\) and a decoder is used to recover the data \(\hat{x} = f_d(z)\). To train the AE, the basic loss function cam be a simple mean squared error \(\mathcal{L}_{MSE} = \|x - \hat{x}\|_2^2\). This basic setting is useful, but have a set of problems, including blurry recovered image due to the MSE function, and a unstructured latent space, where there is no method to direct sample in its latent space and generate new sample by the decoder.

Variational AutoEncoder (VAE) \cite{kingma2013auto}, impose a probabilistic structure on the latent space. The encoder in VAE map the input data into a distribution over the latent space, parameterized by mean \(\mu\) and variance \(\sigma^2\), from which a latent representation \(z\) in sampled. Then the decoder can recover the data using the sampled latent vector. The loss term for VAE, in addition to the MSE loss, has another regularization term to ensure the distribution of the latent space is close to a multivariate Gaussian distribution, usually Kullback-Leibler (KL) divergence chosen. Essentially, the encoder network parameterises a posterior distribution \(q(z|x)\) over the latent variables conditioned on the input data, the loss of KL divergence ensures that this posterior approximates the prior distribution \(p(z)\), and the decoder models the conditional distribution \(p(x|z)\). By enforcing a Gaussian structure in the latent space, VAEs facilitate the interpolation and generation of new data samples that are coherent and varied, enhancing the model's utility in generative tasks. The incorporation of the KL divergence term provides a regularizing effect, making VAEs generally more robust to overfitting compared to traditional autoencoders. However, The Gaussian assumption and the MSE loss often lead to blurring in the reconstructed outputs. This is particularly evident in applications like image reconstruction where sharpness and detail are crucial. In some cases, VAEs can experience mode collapse, where the model ignores certain modes of the data distribution, leading to less diversity in the generated samples.

The Vector Quantized Variational AutoEncoder (VQ-VAE), following the variational autoencoder idea, uses a discrete latent representation. In the VQ-VAE, after getting a code from the encoder, it is seen as a set of feature vectors, which are going to be replaced by their nearest neighbour in a predefined codebook, a fixed collection of feature vectors. This quantization step replaces the Gaussian sampling process seen in standard VAEs, thereby making the latent space discrete. The decoder uses this quantized vector to reconstruct the input data, as is shown in Figure \ref{fig:vqvae-net}.

The loss function for VQ-VAE is still similar to VAE, including a reconstruction loss term, typically MSE. For the regularization term for the latent space, the loss encourage the encoder's output to be close to the selected codebook vector, and also encourage the codebook vector be closer to the encoded feature, which can be written as \cite{van2017neural}:
\begin{align}
    \mathcal{L}_{VQ} = \|sg[f_e(x)]-e\|_2^2 + \beta \|f_e(x)-sg[e]\|_2^2
\end{align}
where \(sg[\cdot]\) means stop gradient. These two terms essentially bring the codebook and encoded vector close together. The prior distribution of the latent space is assumed to be uniform, and the code is directly queried by nearest vector not sampled, the KL divergence is constant w.r.t. the encoder parameters. As a result, the KL divergence term is ignored in VQ-VAE.

By using discrete codebook vectors, VQ-VAE often yields sharper reconstructions than traditional VAE. Furthermore, VQ-VAEs can efficiently encode information, as each vector in the codebook can represent a large and complex pattern within the data, making them particularly useful in tasks like speech and image synthesis where high fidelity is crucial. In SDFusion, the VQ-VAE is adapted to process 3D tensor for object representation instead of 2D, the 2D convolusion operation involved is replaced by 3D.

\subsection{Latent Diffusion} 
\begin{figure}
    \centering
    \includegraphics[width=0.8\linewidth]{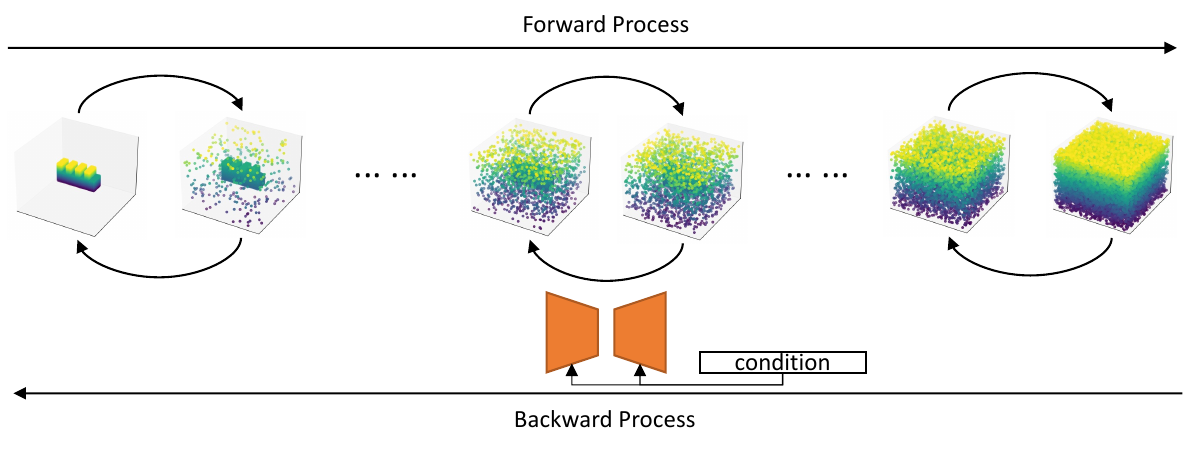}
    \caption{Diffusion process for a voxel represented shape. the forward process is to constantly add Gaussian noise, and the reverse process is learned by training a denoiser. In this report, the true diffusion process happens in latent space instead of voxel space, this is only for illustration of diffusion process. The original idea is from \cite{ho2020denoising}.}
    \label{fig:diffusion}
\end{figure}
Traditional diffusion models \cite{sohl2015deep} are probabilistic models that learn a transformation from a prior Gaussian distribution to the target data distribution \(p(x)\). This transformation is implemented through a fixed Markov Chain, divided into a forward process and a reverse process. In the forward process, Gaussian noise is gradually added to the data, formalized as \(q(x_t | x_{t-1}) := \mathcal{N}(x_{t};\sqrt{1-\beta_t}x_{t-1},\beta_t \mathbf{I})\), where \(\beta_t\) represents a variance schedule across the steps from 1 to a maximum time step \(T\). A denoising function \(\epsilon_\theta(x_t, t)\) is trained to reverse this noising process, reconstructing the original data from the noisy data at each step.

Latent Diffusion Model (LDM) \cite{rombach2022high} modified this approach by targeting the diffusion process in a learned latent space rather than the original pixel space. This shift entails encoding the high-dimensional data, such as images or 3D objects, into a more compact, lower-dimensional latent representation using an autoencoder, which in both LDM \cite{rombach2022high} and SDFusion \cite{cheng2022sdfusion}, VQ-VAE is chosen. The diffusion process then applies to these latent representations, which encapsulates the essential features and characteristics of the data more efficiently than operating directly in pixel space. By reducing the dimensionality of the data, LDM requires fewer computational resources, making the training and sampling processes faster and less memory-intensive. Moreover, diffusing in latent space helps focus the model on capturing and reconstructing the underlying semantic and structural aspects of the data, often leading to higher quality samples that maintain fidelity to the essential attributes of the original dataset.

\begin{figure}
    \centering
    \includegraphics[width=0.8\linewidth]{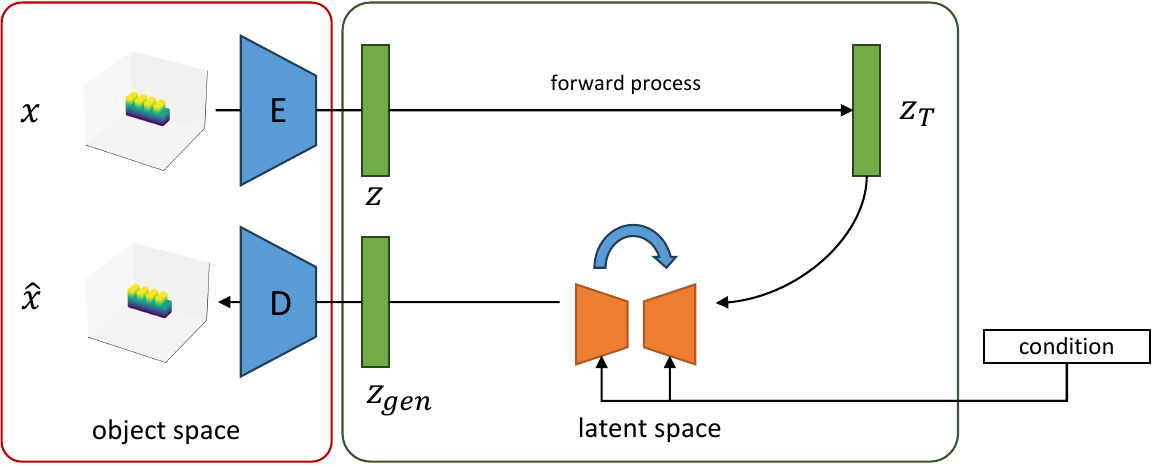}
    \caption{Structure of latent diffusion model \cite{rombach2022high}.}
    \label{fig:ldm}
\end{figure}

\section{Experimental Setup}
\label{sec:experiment}

\subsection{Dataset Collection}

\begin{figure} [htbp]
    \centering
    \includegraphics[width=1\linewidth]{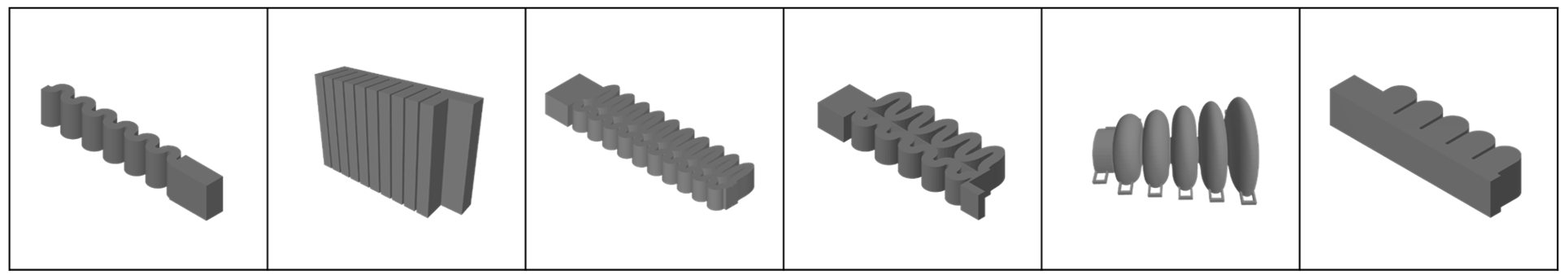}
    \caption{Soft pneumatic robot actuator designs used for dataset.}
    \label{fig:dataset}
\end{figure}
\begin{figure} [htbp]
    \centering
    \includegraphics[width=1\linewidth]{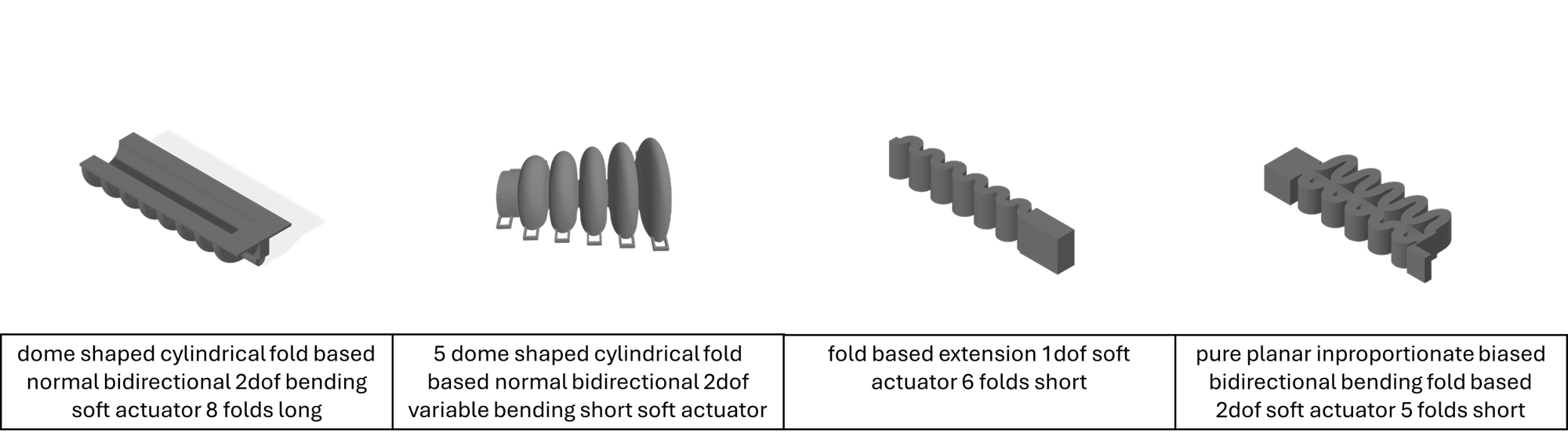}
    \caption{Text-Shape Pairings of Soft Actuators.}
    \label{fig:text-shape-pairing}
\end{figure}
The dataset utilized in this study is focused on soft pneumatic robot actuator 3D models, specifically tailored for training 3D diffusion models. Due to the limited size of the original dataset, we employed data augmentation techniques to expand the total number of training samples. The data augmentation process involves leveraging SolidWorks, a powerful CAD software, to create new shapes with varying parameters. Specifically, global variables within SolidWorks are manipulated to modify key design parameters such as the number of folds, degrees of freedom, and other geometric properties of the soft pneumatic actuators. By systematically varying these parameters, a diverse set of new shapes is generated, effectively expanding the dataset and enriching it with a broader spectrum of soft pneumatic actuator configurations.

To construct the text-shape dataset, a meticulous manual process was undertaken to describe the parameters and characteristics of the soft pneumatic robot actuator designs. These descriptions encompassed various crucial aspects of the actuator's structure, such as its overall shape, number of folds, degree of freedom, and specific mechanical properties like bending, extending, and twisting capabilities. Each text description was meticulously crafted to provide detailed insights into the physical attributes and functionalities of the soft pneumatic actuators. Overall, the dataset consists of more than 70 text-shape pairings, meticulously curated and structured to facilitate the training of the 3D diffusion model. These pairings serve as the foundational data for the model, enabling it to learn and infer the intricate relationships between text descriptions and corresponding 3D shapes.

\section{Results and discussion}
\label{sec:results}

\subsection{Metrics}
\subsubsection{Fréchet Inception Distance}
\textbf{Rendered Fréchet Inception Distance.} The Fréchet Inception Distance (FID) is an established metric widely utilized in the evaluation of image generation models. It involves extracting features from images using the Inception v3 architecture, treating these features as elements in a multidimensional Gaussian distribution. The FID is calculated by comparing the statistical properties, specifically, the mean \(
\mu\) and covariance \(\Sigma\), of the feature distributions from real and generated images. The distance formula is given by:

\begin{align}
\label{eq:FID}
    \text{FID}(r, g) = \|\mu_r + \mu_g\|_2^2 + \text{Tr}(\sqrt{\Sigma_r+\Sigma_g-2\Sigma_r\Sigma_g})
\end{align}

\((\mu_r,\Sigma_r)\) and \((\mu_g,\Sigma_g)\) are mean and covariance of the real and generated data, respectively. With lower FID, means these two distribution more similar. FID can both evaluate the fidelity and the diversity. To compare the shape level similarity, for each generated shape, a set of 360 degree render is generated both from above and below the shape to provide a full view. With the generated set and ground truth set of image, FID is calculated accordingly.

\textbf{3D Fréchet Distance.} Similar to FID, we introduced the 3D Fréchet Distance for direct shape analysis. This adaptation uses a 3D-specific encoder rather than image-based Inception v3, enabling the direct evaluation of 3D shapes. By encoding the geometric data directly, the 3D Fréchet Distance offers a novel approach to measuring the fidelity of 3D models, ensuring that the shapes are not only visually accurate but also geometrically precise.

For the 3D encoder we use the encoded result before quantization with VQVAE pretrained on ShapeNet only. The encoded tensor in VQVAE is shape like \texttt{[3, 16, 16, 16]} with a total 12288 features, which is too high dimension to calculate the covariance matrix. As a result, an \({2^3}\) 3D average pooling is used to get a \texttt{[3, 8, 8, 8]} feature vector. Then the Fréchet Distance is calculated the same as (\ref{eq:FID}).

\subsubsection{Surface Smoothness}
In the evaluation of 3D model quality, surface smoothness is a critical factor, reflecting the quality of the generation of the representation. To quantify this aspect, we employ the standard deviation of normal variation as our primary metric. This statistic measures the angular differences between normals of adjacent faces, providing a gauge of the geometric consistency across a model’s surface. A lower standard deviation indicates a smoother surface, signifying fewer abrupt changes in angle, which are often perceptible as visual and tactile roughness in physical models. We found that this metric have a good alignment with human perception. One limitation it presents is that this can be only comparable among the generated models, as they share a similar mesh structure, with a dense triangular mesh. While for the Computer Aided Design Software (CAD) generated \texttt{.obj} file, they face count are much less, typically with one or two normal per face, leading to a much larger variation if compared with the dense mesh.

\subsubsection{Volumetric Intersection over Union}
The Volumetric Intersection over Union (IoU) metric extends the traditional IoU metric to evaluate the accuracy of volumetric reconstructions in three-dimensional spaces. It is defined mathematically as:
\[
\text{Volumetric IoU} = \frac{\text{Volume of Intersection}(\mathcal{V}_{pred}, \mathcal{V}_{gt})}{\text{Volume of Union}(\mathcal{V}_{pred}, \mathcal{V}_{gt})}
\]
where $\mathcal{V}_{pred}$ represents the predicted 3D volume and $\mathcal{V}_{gt}$ denotes the ground truth 3D volume. The \textit{Volume of Intersection} is calculated by counting the voxels that are occupied in both the predicted and the ground truth volumes. The \textit{Volume of Union} includes all voxels occupied in either of the volumes, encompassing the intersection.

IoU was first used as a metrics for object detection. Tatarchenko et al. \cite{tatarchenko2019single} first use it for 3D reconstruction.

\subsection{VQ-VAE}

The primary challenge we encountered when training the VQ-VAE was the limited size of our dataset. ShapeNet, for instance, contains over 300 million objects, including 236 thousand chairs. Such a small dataset is inadequate and can result in poor performance. To address this issue, we employed transfer learning and data augmentation techniques.

\begin{figure}
    \centering
    \includegraphics[width=0.8\linewidth]{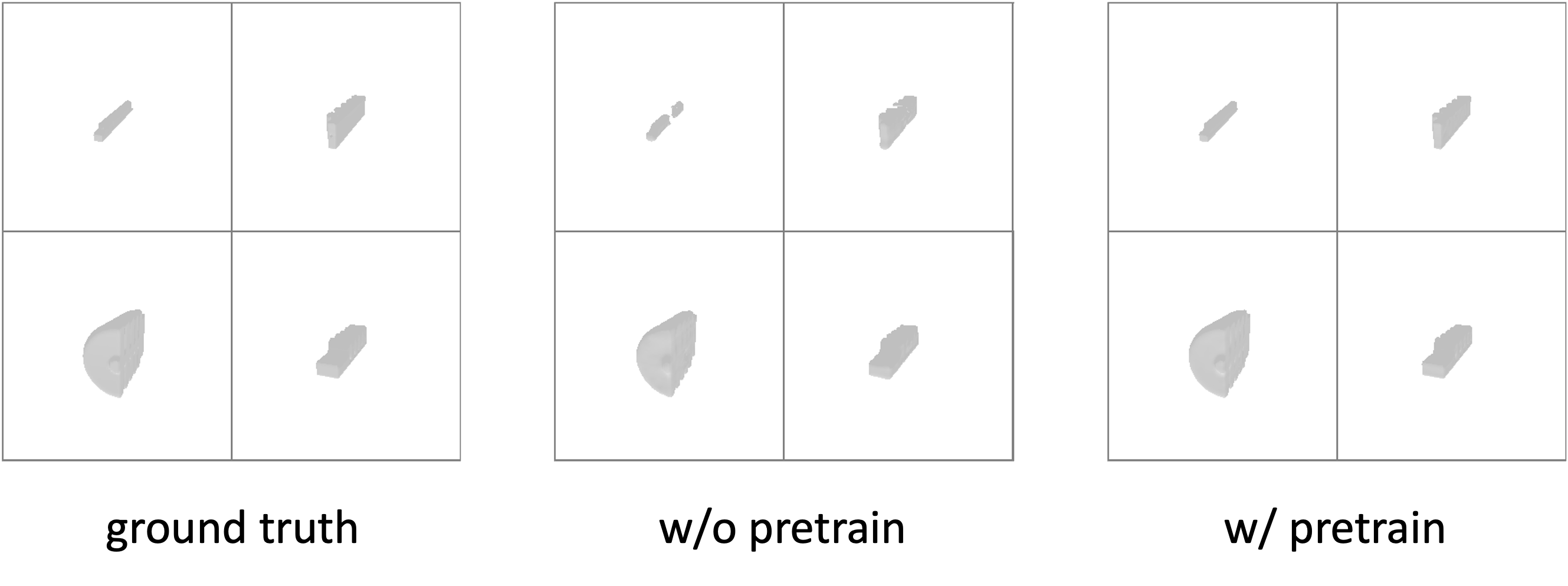}
    \caption{Qualitative results of VQ-VAE with and without pretraining. Both results are finally trained on the augmented dataset.}
    \label{fig:res_vqvae}
\end{figure}

The quantitative results of the VQ-VAE are presented in Table \ref{tab:comp_vqvae}. In image representation learning, it is often assumed that an encoder trained on a large, diverse dataset is sufficient. However, testing revealed that a VQ-VAE trained solely on ShapeNet struggled to perform well in our scenario, likely due to differences in the shapes of actuators compared to everyday objects in ShapeNet. As shown in the table, with the implementation of these two improvements, the results significantly improved. It is worth noting that only results with an Intersection over Union (IoU) above approximately 0.75 are considered acceptable. The qualitative results are shown in Figure \ref{fig:res_vqvae}. Following these enhancements, the VQ-VAE was able to achieve faithful reconstructions, which is crucial for subsequent training stages.

\begin{table}
    \centering
    \begin{tabular}{lcc}
        \toprule
        metric & \({L}_{recon}\) & IoU \\
        \midrule
        trained on ShapeNet & 3.97e-3 &0.57 \\
        trained on SoftNet  & 7.67e-3 &0.21 \\
        + Transfer Learning & 1.13e-3 &0.61 \\
        + Data Augmentation & \textbf{2.80e-4} & \textbf{0.89} \\
        \bottomrule
        
    \end{tabular}
    \caption{Comparison experiments of VQVAE}
    \label{tab:comp_vqvae}
\end{table}

\subsection{Text to Shape}

\begin{figure} [htbp]
    \centering
    \includegraphics[width=0.8\linewidth]{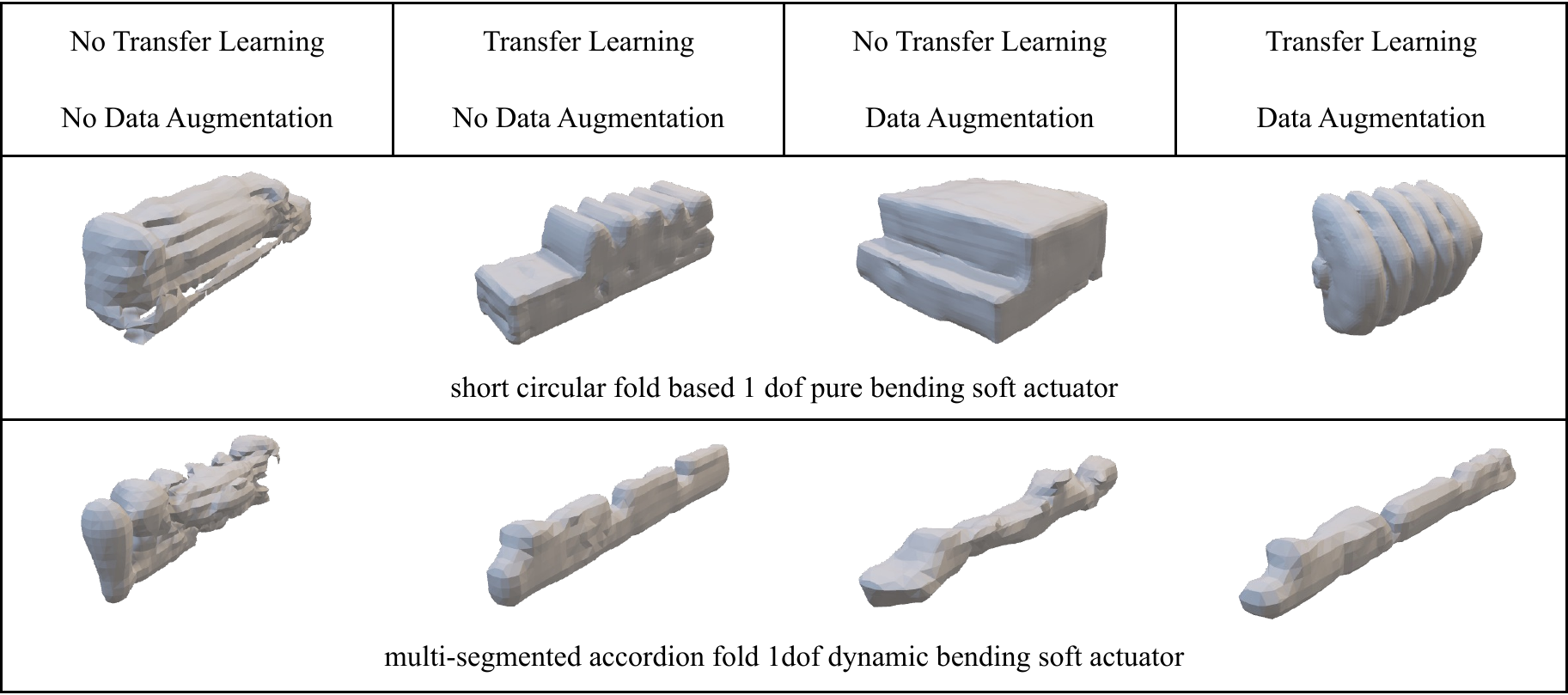}
    \caption{Qualitative results of text to shape generation with or without transfer learning and data augmentation. Is is shown that transfer learning can largely help the surface quality and the shape, while data augmentation can improve the understanding of the shape.}
    \label{fig:quali-comp}
\end{figure}

\begin{figure} [htbp]
    \centering
    \includegraphics[width=0.8\linewidth]{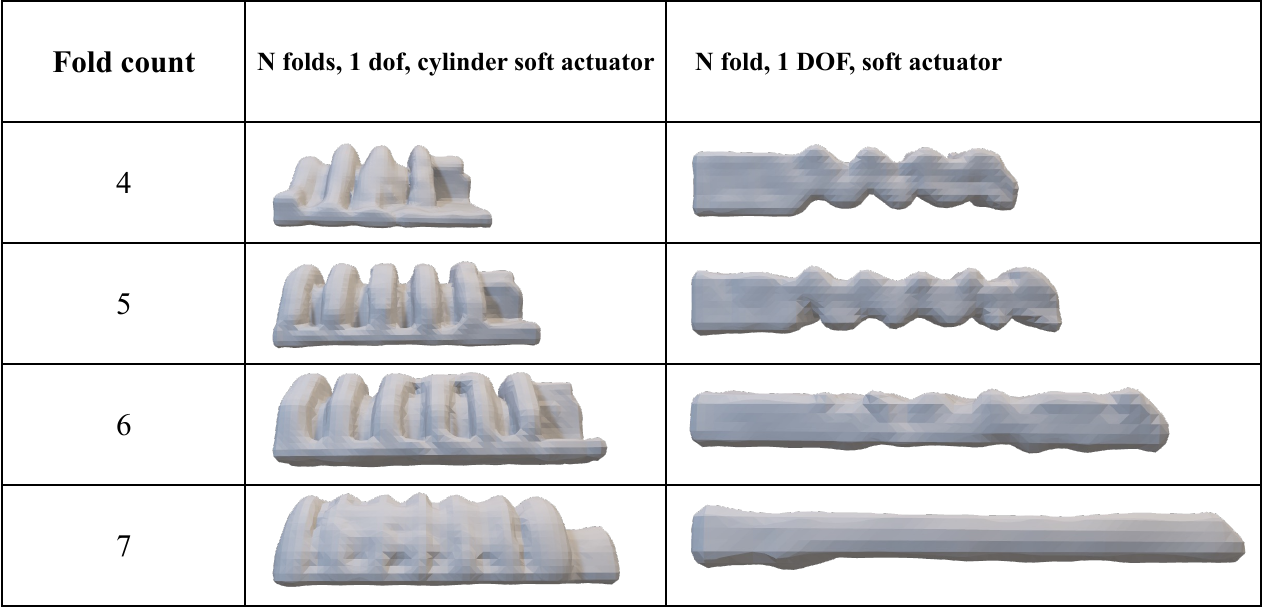}
    \caption{Qualitative results of how the text to shape generation respond to different number of folds. The model can understand the description with data augmentation. The lost of detail in long actuator is limited with the target resolution.}
    \label{fig:res-folds}
\end{figure}

To evaluate the effectiveness of the diffusion model, we generated 100 lines of text prompts similar to those in the training dataset. The qualitative comparison is presented in Figure \ref{fig:quali-comp}. Regarding shape completeness, the results align with the surface quality metric. Direct training yielded the poorest results, while transfer learning showed significant improvement. To assess the models' understanding of the prompts, we present two examples here and generate results from each model. For the first prompt, only the final model produced the correct generation. For the second prompt, three models were close to the correct generation, but the final one was the most accurate. However, all models lack important details, likely due to limitations in resolution. We also investigated the models' ability to respond to different numbers of folds, as shown in Figure \ref{fig:res-folds}. In some cases, the models performed well, but as the complexity of the model increased, fewer details were visible due to limitations in total resolution.

\begin{table}
    \centering
    \begin{tabular}{lcccc}
        \toprule
         Data Augmentation&  no&   no&yes&  yes\\
         Transfer Learning&  no&   yes&no&  yes\\
         \midrule
         Normal std&  0.45&   0.19&0.26&  \textbf{0.18}\\
         RFID (gen)& 225.22&  127.56& 198.03&\textbf{113.06}\\
         3D FD& 24.15& 23.68& 22.00&\textbf{20.23}\\
    \bottomrule
    \end{tabular}
    \caption{Comparison experiments of Latent Diffusion Results}
    \label{tab:comp_txt2shape}
\end{table}

In conclusion, transfer learning plays a crucial role in helping the model achieve better mesh and surface quality. A larger dataset will also aid the model in understanding the inner relations of different parameters. However, a limitation is the small resolution, which is constrained by GPU performance. The quality could potentially be higher with the same method but with a larger dataset. Currently, the generation process only considers shape, without taking into account physics and materials. Therefore, there are several compromises in the current approach. Nevertheless, scaling up should significantly improve the results.

\section{Conclusion and future work}

In this report, we explore the generation of soft robot designs using generative AI. We collect and augment pneumatic soft robot designs to form a dataset. Then, we adapt a latent diffusion model to learn the data distribution and generate novel designs from it. The results are promising but are largely limited by current GPU performance and dataset size. We believe that with more GPUs and a better structure assignment, such as separating VQ-VAE onto a dedicated GPU, we can boost performance and scale up the resolution, which is crucial for generating potentially printable designs. A larger dataset is clearly advantageous, but this limitation can be alleviated by combining it with simulation. Without this, thousands of models would benefit greatly. In the current setup, only shape is considered, and no physics simulations are involved. However, this method could potentially be combined with simulation through reinforcement learning. This would allow the model to benefit from mimicking current human design while also learning from the physical world.




\section*{Acknowledgments}
The authors gratefully acknowledge the invaluable contribution of Benjamin WK Ang, whose soft robot design served as the foundation for constructing the dataset used in this research project.

\bibliographystyle{unsrt}  
\bibliography{references}

\end{document}